\documentclass[runningheads]{llncs}
\usepackage[T1]{fontenc}
\usepackage[numbers,sort&compress]{natbib}
\usepackage{alltt}
\usepackage[breakable,skins]{tcolorbox}
\usepackage{amsmath,amssymb,amsfonts}
\usepackage{algorithmic}
\usepackage{graphicx}
\usepackage{float}
\usepackage{subcaption}
\usepackage{textcomp}
\usepackage{xcolor}
\usepackage{comment}
\usepackage{booktabs}
\usepackage{url}
\usepackage{tabularx}
\usepackage{csquotes}
\tcbset{
  aibox/.style={
    width=345.18663pt,
    top=10pt,
    colback=black!05,
    colframe=black!20,
    colbacktitle=black!50,
    enhanced,
    center,
    attach boxed title to top left={yshift=-0.1in,xshift=0.15in},
    boxed title style={boxrule=0pt,colframe=white,},
  },
  innerbox/.style={
    sharp corners,
    colframe=black!60, 
    boxrule=0.5pt,  
    colback=white,  
    colbacktitle=black!50,
    enhanced,
    boxsep=-2pt,     
  }
}
\newtcolorbox{AIBox}[2][]{aibox,title=#2,#1}
\newtcolorbox{InnerBox}[1][]{innerbox,#1}

\title{LoopBench: Discovering Emergent Symmetry Breaking Strategies with LLM Swarms}

\author{Ali Parsaee\inst{1} \and Yashar Talebirad\inst{1} \and Csongor Szepesvári\inst{1} \and Vishwajeet Ohal\inst{1} \and Eden Redman\inst{2}}

\authorrunning{A. Parsaee et al.}

\institute{University of Alberta, Edmonton, Canada\\
\email{\{parsaee,talebira,csongor,ohal\}@ualberta.ca}
\and
Network for Applied Technology, Edmonton, Canada\\
\email{eden@nat.ltd}}

\begin{document}

\maketitle

\begin{abstract}
Large Language Models (LLMs) are increasingly being utilized as autonomous agents, yet their ability to coordinate in distributed systems remains poorly understood. We introduce \textbf{LoopBench}, a benchmark to evaluate LLM reasoning in distributed symmetry breaking and meta-cognitive thinking. The benchmark focuses on coloring odd cycle graphs ($C_3, C_5, C_{11}$) with limited colors, where deterministic, non-communicating agents fail in infinite loops. A strategy passing mechanism is implemented as a form of consistent memory.  We show that while standard LLMs and classical heuristics struggle, advanced reasoning models (e.g., O3) devise strategies to escape deadlocks. LoopBench allows the study of emergent distributed algorithms based on language-based reasoning, offering a testbed for collective intelligence.

\end{abstract} 

%%%%%%%%%%%%%%%%%%%%%%%%%%%%%%%%%%%%%%%%%%%%%%%%%%%%%%%%%%%%%%%%%%%%%%%%

\section{Introduction}
Large Language Models (LLMs) are evolving beyond isolated chatbots into the building blocks of autonomous multi-agent systems. The true potential of these "AI swarms" lies not just in individual problem-solving, but in their ability to solve complex problems without a central coordinator. However, coordinating a distributed system of reasoning agents presents a fundamental challenge: can independent LLMs, driven by local prompts and limited observations, autonomously invent the strategies necessary to collaborate? This ability hinges on \emph{meta-cognitive reasoning}, which we define as the capacity to adopt strategies that go beyond immediate local optimization.

We wanted to test for meta-cognitive reasoning by creating a setting where thinking narrowly results in deadlock, but 'zooming out' reveals the solution. This led us to the fundamental problem of loop breaking in graphs. Specifically, we focus on over-constrained odd cycles, a scenario where perfect solutions are impossible and simple greedy heuristics lead to endless oscillation loops and fail. This setting forces agents to go beyond immediate conflict resolution and adopt novel strategies that benefit the collective instead.

We introduce \textbf{LoopBench}, a benchmark focused on symmetry breaking in over-constrained odd-cycle graphs ($C_3, C_5, C_{11}$). By analyzing quantitative performance (conflict minimization) and qualitative strategy evolution (via feed-forward strategies), we measure the "Reasoning Gap" between models.

We demonstrate that advanced reasoning models like O3 successfully break symmetries by developing strategies such as "waiting" or history-based pseudo-randomness. We interpret this capacity to detect and escape deadlock as a form of meta-cognitive thinking, enabling agents to override immediate greedy incentives for long-term coordination. In contrast, classical heuristics and weaker LLMs often remain stuck.
This work explores the following:

\begin{enumerate}
    \item A framework for evaluating distributed agents on symmetry-breaking tasks.
    \item An LLM-agent architecture using consistent memory to enable emergent and evolving strategies.
    \item Experiments showing a performance gap between different LLMs as well as algorithmic baselines.
    \item Qualitative analysis of emergent strategies showing how meta-cognitive reasoning arises from local observations.
\end{enumerate}

\section{Related Work}
\label{sec:related}

General multi-agent frameworks like \textit{AgentVerse} \citep{chen2023_agentverse} and \textit{AgentBench} \citep{liu2023_agentbench} investigate how LLM agents collaborate within structured environments. \textit{GPTSwarm} \citep{zhuge2024_gptswarm} takes a graph-optimization approach, refining both agent prompts and their connectivity. For graph-based tasks specifically, \textit{GraphAgent} \citep{hu2024_graphagent} employs a distributed approach where node-specific agents communicate synchronously, achieving high accuracy on large-scale polynomial-time problems. The \textit{AgentsNet} benchmark \citep{agentsnet2025} evaluates decentralized protocols like leader election and coloring under strict communication constraints. 

Self-improvement mechanisms have also proven effective. \emph{Reflexion} \citep{shinn2023_reflexion} uses verbal reinforcement and episodic memory to help agents correct mistakes in subsequent trials. Research on self-reflection indicates that structured templates, such as prompts for retrying or summarizing solutions, enhance performance across various domains \citep{renze2024_selfreflection}. Additionally, \emph{FINEREASON} \citep{chen2025_finereason} Demonstrates that practicing self-correction in puzzles improves mathematical reasoning.

LoopBench aligns with multi-agent reasoning and reflective frameworks, but instead of introducing a new solver, we provide a \textbf{minimal benchmark} designed to test symmetry breaking under severe information constraints. Our feed-forward strategy architecture, while inspired by self-reflection, allows us to trace the emergence of novel strategies, facilitating a direct comparison between models with varying reasoning capabilities.

\section{LoopBench: A Benchmark for Symmetry Breaking}

Let $G=(V,E)$ be an undirected graph with $|V|=n$ vertices and $|E|=m$ edges.  
A \emph{$c$-coloring} is a function $C:V\!\to\![c]$ that assigns one of $c$ available colors to each vertex, where $[c]=\{0,1,\cdots,c-1\}$.  
An edge $\{u,v\}\!\in\!E$ is said to be in \emph{conflict} if $C(u)=C(v)$.  
The corresponding conflict set is $E(C)=\{\{u,v\}\!\in\!E : C(u)=C(v)\}$.

Symmetry breaking is a fundamental challenge in distributed systems. In a perfectly symmetric environment, deterministic agents executioning the same algorithm will oscillate. As an example, for an odd cycle $G=C_n$, where every node is initialized with the same color between $c=2$ colors, conflicts are unavoidable. If nodes share a color, greedy rules often cause simultaneous switching, perpetuating deadlock. Breaking such symmetries requires randomness or unique identifiers, making odd cycles a great testbed for evaluating an agent's ability to think and strategize beyond greedy algorithms. 

Let $\chi(G)$ denote the chromatic number. We call a graph \textbf{Over-constrained} if $c < \chi(G)$. The goal in \emph{soft graph coloring} \citep{Fitzpatrick2002SoftColouring} is to minimize conflicts, balancing constraint satisfaction against feasibility.

The goal in \emph{soft graph coloring} \citep{Fitzpatrick2002SoftColouring} is not necessarily to eliminate all conflicts (which are impossible in over-constrained settings), but rather to minimize them.
The optimization task for all settings is thus: minimizing the number of conflicts,
subject to each vertex operating with only local neighborhood information.  
This forms the foundation for comparing classical distributed heuristics and LLM-based agent behavior.

\paragraph{Evaluation Metrics.}
To compare agent performance, we report two metrics computed over $T$ steps, normalized relative to the theoretical minimum possible conflicts ($\text{conf}_{best}$) for a given graph (e.g., 1 for odd cycles with 2 colors).

\begin{enumerate}
    \item \textbf{Proximity to Optimal (\%):} Measures how close the agent's conflict count is to the theoretical best on average.
    \[
    \text{Proximity} = 100 - \left( \frac{1}{T} \sum_{t=1}^{T} \frac{\text{conf}_t - \text{conf}_{best}}{\text{conf}_{initial} - \text{conf}_{best}} \right) \times 100
    \]
    A score of 100\% indicates the agent stayed at the theoretical optimum for the entire run. 0\% corresponds to remaining at the initial random state.

    \item \textbf{Stability (\%):} Measures the agent's ability to avoid worsening the conflict state (i.e., preventing regression).
    \[
    \text{Stability} = 100 - \left( \frac{\sum_{t=1}^{T-1} \mathbb{I}(\text{conf}_{t+1} > \text{conf}_t)}{T-1} \right) \times 100
    \]
    where $\mathbb{I}$ is the indicator function. High stability means the agent rarely makes moves that increase the number of conflicts.
\end{enumerate}

\section{Methodology}
\label{sec:methodology}

We evaluate distributed symmetry breaking on Odd Cycle Graphs ($C_3, C_5, C_{11}$) constrained to $c=2$ colors. This over-constrained setting forces agents to coordinate to minimize conflicts rather than solving for a perfect zero-conflict state. We instantiate each vertex $v \in V$ as an independent agent in a synchronous execution model. We compare LLM-based agents against four baselines: \emph{Soft Colorer FP} (updates with $p=0.3$), \emph{Soft Colorer CFP} (only conflicted updates), \emph{Conservative Random}, and \emph{Random}. We exclude deterministic greedy algorithms as they provably fail in symmetric cycles.

At each round $t$, every LLM agent receives a textual prompt containing its local interaction history $\mathcal{H}_t(v)$ which enables the detection of oscillations, where $\mathcal{H}_t(v)$ is
defined as: $$\mathcal{H}_t(v) = (\{C_{t-1}(v), C_{t-1}(v_{adj})\}, \ldots, \{C_0(v), C_0(v_{adj})\})$$
Each agent outputs a color choice $C_t(v) \in [c]$ using OpenAI's structured outputs (Figure~\ref{fig:prompt_full}). To enable strategy evolution, we implement a \textbf{feed-forward} mechanism where agents write short private notes summarizing their reasoning, which are re-injected into the next prompt as consistent memory. We ran the experiments for 15 steps for each configuration, and repeated each experiment setting 5 times.

\begin{figure*}[!t]
\begin{AIBox}{LLM Agent Prompt Template (With Feed-Forward Notes)}
{
\parbox[t]{0.48\textwidth}{{\small \bf System Prompt:} \scriptsize\begin{alltt}
You are a node in a graph coloring
problem. Your goal is to achieve a
stable global coloring where no
neighbors share the same color.
\end{alltt}
\vspace{6pt}
{\small \bf API Parameters:}
\scriptsize\begin{alltt}
- Model: e.g., OpenAI GPT series, O3 series
- Temperature: 1.0
- Response format: json\_schema (strict)
\end{alltt}}%
\parbox[h]{0.04\textwidth}{\hspace{1pt}}
\parbox[t]{0.48\textwidth}{{\small \bf Output JSON Schema:} \scriptsize\begin{alltt}
\{ "type": "object",
  "properties": \{
    "color": \{ "type": "integer",
               "enum": [0, 1, ..., c-1] \},
    "strategy": \{ "type": "string" \} \},
  "required": ["color", "strategy"],
  "additionalProperties": false \}
\end{alltt}
\vspace{6pt}
{\small \bf Example Output:} \scriptsize\begin{alltt}
\{ "color": 1,
  "strategy": "[MODIFIED] Prioritize
colors not used by neighbors to
minimize conflicts. [SAME] Track
historical success rates of each
color. [NEW] Wait one turn before
switching if conflicts decrease." \}
\end{alltt}}

\vspace{8pt}
\noindent\rule{\textwidth}{0.4pt}
\vspace{4pt}

\noindent{\small \bf User Prompt Template:} \scriptsize\begin{alltt}
You are a node in a graph coloring problem. Your task is to choose a color
that minimizes conflicts in the global graph (but you can only observe your
immediate neighbors). Consider all information provided below to develop a
strategy and choose a color.
\#\#\# LOCAL INFORMATION: - Your current color: \textit{<integer>} - Neighbors'
colors: \textit{<node\_id: color dict>} - Available colors: [0, 1, ..., c-1] -
Colors currently used by neighbors: \textit{<list of colors>}
\#\#\# STRUCTURAL INFORMATION: - Your node ID: \textit{<integer>} - Number of
neighbors (node degree): \textit{<integer>} - Neighbor IDs: \textit{<list of node IDs>}
\#\#\# HISTORICAL DATA: - Your color history: \textit{<list of past colors>} -
Your conflict history: \textit{<list of past conflict counts>} - Neighbors'
color history: \textit{<node\_id: color history dict>} - Color performance
history: \textit{<color: performance stats dict>}
\#\#\# PERFORMANCE FEEDBACK: - Current conflict count: \textit{<n>} neighbors
share your color - Recent conflict rate: \textit{<rate>} conflicts per turn -
Current success status: \textit{<CONFLICT FREE | n CONFLICTS>}
\#\#\# MY PRIVATE NOTES: \textit{<previous strategy text from last round>}
Based on your experience, update your private notes with concise, transferable
insights. Focus on documenting general patterns and strategies that have
emerged from observing neighbor behavior and conflict resolution, not specific
color choices. Your notes should be a single string containing a list of
strategies written in a concise, pseudocode-like style. When updating, follow
these rules: 1. Prefix each strategy with `[NEW]`, `[MODIFIED]`, or `[SAME]`.
2. If modifying, keep the original wording as much as possible. 3. Use a
newline to separate different strategies.
\end{alltt}
}
\end{AIBox}
\caption{LLM agent prompt template and output schema with feed-forward notes.}
\label{fig:prompt_full}
\end{figure*}

\section{Results and Discussion}

\begin{table*}[t]
\centering
\caption{Performance on Odd Cycles (Higher is Better). Metrics computed over the full trajectory.}
\label{tab:cycle_results}
\footnotesize
\begin{tabular}{lcccccc}
\toprule
Agent & \multicolumn{2}{c}{\textbf{Graph C3}} & \multicolumn{2}{c}{\textbf{Graph C5}} & \multicolumn{2}{c}{\textbf{Graph C11}} \\
 & \shortstack{Proximity\\(\%)} & \shortstack{Stability\\(\%)} & \shortstack{Proximity\\(\%)} & \shortstack{Stability\\(\%)} & \shortstack{Proximity\\(\%)} & \shortstack{Stability\\(\%)} \\
\midrule
O3 & 72.5 & \textbf{98.7} & 57.5 & \textbf{100.0} & 69.9 & \textbf{99.0} \\
GPT-4.1 Nano & 1.2 & 98.7 & 4.4* & 92.0 & 2.5* & 89.0 \\
Soft Colorer FP & \textbf{75.0} & 89.3 & \textbf{80.0} & 93.3 & \textbf{83.2} & 93.0 \\
Soft Colorer CFP & 67.5 & 90.7 & 78.8 & 88.0 & 78.9 & 89.0 \\
Conservative Random & 67.5 & 82.7 & 72.5 & 82.7 & 62.3 & 74.0 \\
Random & 68.8 & 86.7 & 54.4 & 85.3 & 50.0 & 66.7 \\
\bottomrule
\end{tabular}
\end{table*}

Table~\ref{tab:cycle_results} summarizes performance across agent types on cycles $C_3$, $C_5$, and $C_{11}$. We report Proximity and Stability over the duration of the simulation. As shown, there is a massive performance gap between reasoning-optimized models (O3) and standard small models (GPT-4.1 Nano). O3 consistently achieves high proximity scores (55--72\%) and breaks symmetry effectively. In contrast, GPT-4.1 Nano fails almost completely, with proximity scores near 0\% on all cycles. We also tested additional models including GPT-4.1, GPT-4.1-mini, and O3-mini, all of which similarly failed to break out of oscillation loops. Across all graphs, O3 achieves near-perfect stability (98--100\%), meaning it almost \emph{never} makes moves that increase conflicts after converging, meaning it is much better than other methods at noticing when the optimal point has been reached. This is significantly higher than classical baselines (78--92\%). 

The Soft Colorer FP baseline performs very well, obtaining the highest proximity scores of all tested models. However, the goal of this work is not to generate the best algorithm for graph coloring, but rather to show LLMs are capable of generating strategies that require meta-cognitive reasoning. We believe if more capable models run for longer steps, or if they are asked to make strategies after being exposed to many different types of graphs, eventually they may come up with strategies competitive to the Soft colorers in proximity. Furthermore, while Soft Colorer relies on a hard-coded probability ($p=0.3$), O3 \emph{learns} to hold its color dynamically. Finally, O3 does outperform the Soft Colorer algorithms in terms of stability, after reaching the optimal 1-conflict state.

\subsection{The Emergence of Strategies}
A key advantage of LLM-based agents is their apparent interpretability. By analyzing the strategies, we can trace the emergence of coordination strategies. We observe that significant performance improvements often coincide with revisions to these notes. Inspection of the O3 agent's notes on the $C_{11}$ graph reveals an evolution from simple reactive rules to more sophisticated, history-aware heuristics. 
Initially, agents often start with a basic greedy strategy like:
\begin{displayquote}
\textit{"[NEW] Prioritize colors not used by neighbors to minimize conflicts."}
\end{displayquote}
After observing oscillations, a "eureka" moment occurs when agents discover waiting strategies, preventing the deadlock: 
\begin{displayquote}
\textit{"[MODIFIED] If in conflict, wait one turn before switching to see if neighbors resolve it."}
\end{displayquote}
 Later, agents develop strategies using historical data: \begin{displayquote}\textit{"[NEW] Track historical success rates of each color and favor those with a lower conflict history."}\end{displayquote}

\begin{figure*}[!t]
    \centering
    \includegraphics[width=\textwidth]{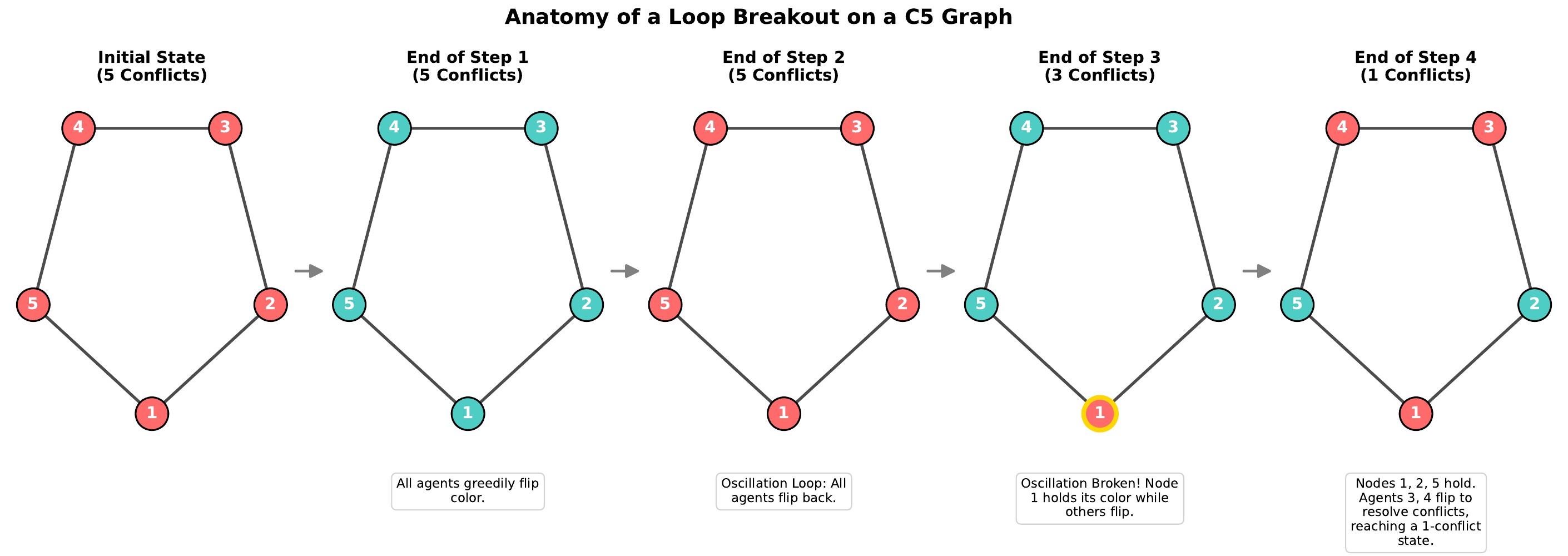}
    \caption{Visualization of a C5 graph coloring process. The system begins in a high-conflict state (a) and enters an oscillation loop (b-c). At Step 3 (d), symmetry is broken as Nodes 1, 2, and 5 hold their colors (using "wait" strategies), while others continue to flip. This partial stabilization allows Agents 3 and 4 to resolve the remaining conflicts in Step 4 (e), reaching the optimal one-conflict state.}
    \label{fig:loop_breakout}
\end{figure*}

\begin{figure*}[!t]
    \centering
    \includegraphics[width=\textwidth]{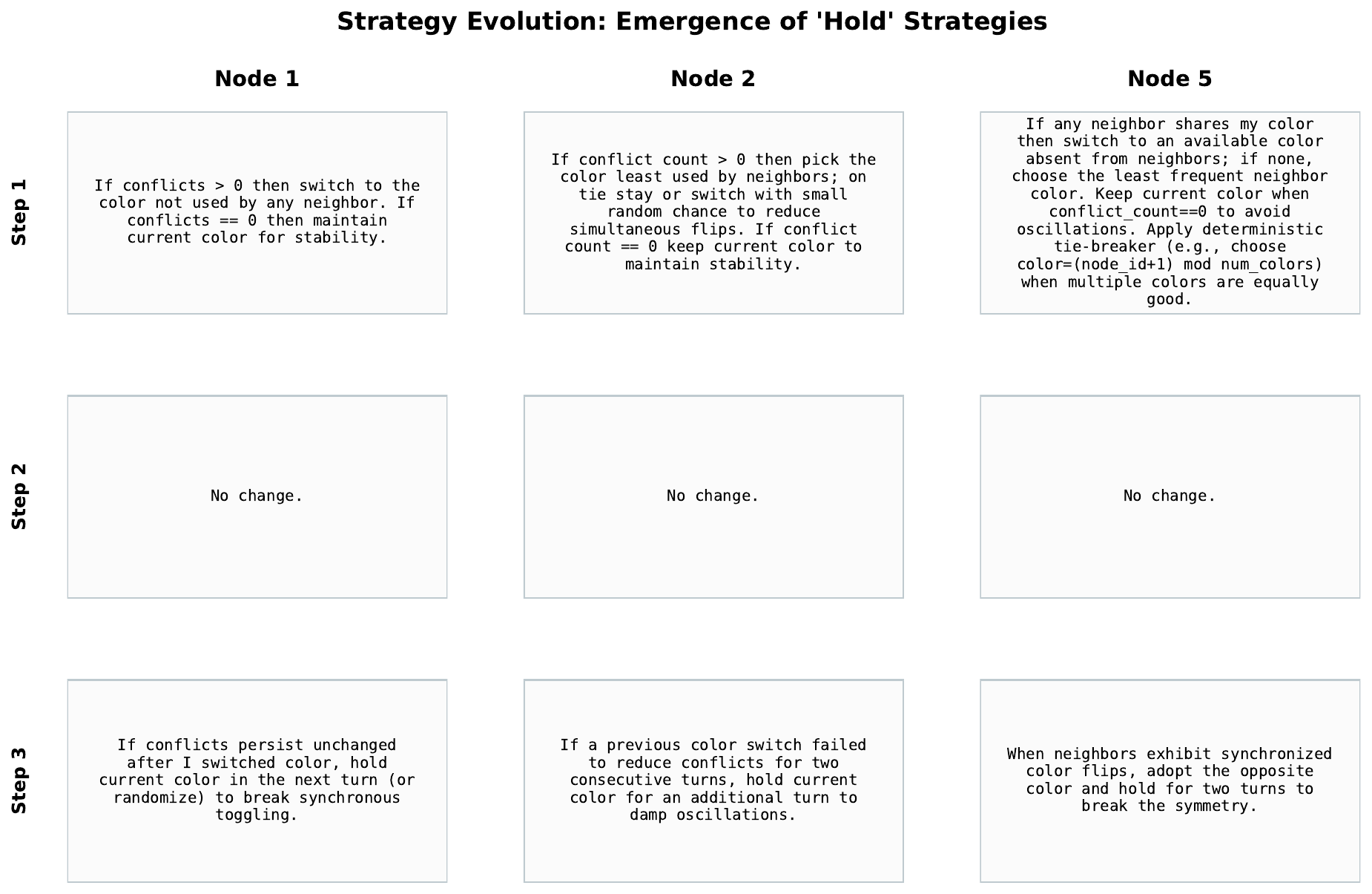}
    \caption{Evolution of internal strategies for Nodes 1, 2, and 5 corresponding to the breakout event in Figure~\ref{fig:loop_breakout}. While all nodes start with greedy heuristics in Step 1, they independently develop "hold" or "dampening" strategies by Step 3. This cognitive shift from reactive flipping to patience allows the system to escape the oscillation loop.}
    \label{fig:strategies_evolution}
\end{figure*}

This evolution shows the agent's ability to reason about the problem dynamics and discover new solutions. By examining the failure of simple greedy heuristics and adopting counter-intuitive behaviors (like waiting), the agents demonstrate meta-cognitive reasoning about their own strategy rather than just the state.  Figure~\ref{fig:loop_breakout} visualizes the breakout of an oscillation loop, while Figure~\ref{fig:strategies_evolution} traces the corresponding evolution of the agents' internal strategies.\\

% To systematically trace how agents refine their heuristics over time, we developed an automated lineage-tracking algorithm that reconstructs the evolutionary tree of each agent's notes over time. Agents annotate their strategy updates with \textit{[NEW]} (novel insights), \textit{[MODIFIED]} (refinements), or \textit{[SAME]} (unchanged heuristics), enabling us to construct parent-child relationships across timesteps. The algorithm operates in three phases: first, strategies tagged \textit{[SAME]} are matched to their predecessors via exact text similarity; second, \textit{[MODIFIED]} entries are paired with the most similar unmatched strategy from the prior timestep using sequence matching (threshold $\geq 0.35$); third, any remaining unmatched entries connect directly to the root, indicating independent emergence. For each modification edge, we compute word-level diffs to identify which concepts were added or removed, extracting \emph{buzzwords}, which are salient vocabulary unique to each revision, that mark conceptual changes. When agents produce identical strategies across consecutive steps, we condense these into step-range nodes (e.g., ``Steps 15--22'' to highlight phases of active exploration versus stable convergence. This algorithmic view into the ``cognitive trajectory'' of individual agents reveals \emph{when} and \emph{how} eureka moments occur, complementing aggregate performance metrics with insight into emergent coordination.

Overall, the experiments suggest that self-generated memory combined with self-reflection enables LLM agents to rediscover classic heuristics and possibly specialize adaptively. Notably, while the classical \emph{Soft Colorer FP} baseline performed well, it relied on a fixed, pre-tuned probability parameter ($p=0.3$). In contrast, LLM agents effectively \emph{auto-tuned} their own update probabilities by dynamically adopting "hold" strategies based on observed conflict patterns, showing a form of meta-learning that may prove more robust in dynamic or unknown environments. However, the diversity in outcomes, ranging from optimal convergence to ineffective overthinking, points to high variance in how effectively reflection translates to action. Furthermore, future work should explore reward shaping to ensure that every ``idea" in the private note reliably translates into improved action.

\subsection{Knowledge Distillation and Frontier Generalization}
We present two sets of preliminary results that point toward promising directions for scalable multi-agent reasoning.

First, a key barrier to deploying reasoning models is cost and latency. We hypothesized that while smaller models (e.g., O3-mini) struggle to \emph{discover} effective strategies from scratch, they might be capable of \emph{executing} them if provided with the right heuristics. To test this, we took the final, evolved strategies from a successful O3 run (containing explicit ``wait'' and ``history-tracking'' heuristics) and injected them into the initial prompt of O3-mini agents. 
In our baseline experiments, O3-mini consistently failed to break symmetries on $C_3$. However, with this ``strategy pre-training,'' the O3-mini swarm successfully broke the oscillation loop in 2 out of 3 trials. This finding bridges the gap between reasoning and execution: while we know LLMs can write code, this demonstrates they can also \emph{act} as the code, executing distributed protocols purely through in-context learning. This suggests a ``Discovery-Implementation Gap'': high-level reasoning is required to generate coordination protocols, but once articulated in natural language, these protocols can be executed by more efficient models. This hints at a novel method for ``distilling'' agentic reasoning into textual heuristics.

Secondly, we also conducted limited experiments with \textbf{Claude 4.5 Sonnet}, \textbf{Gemini 2.5 Pro}, and \textbf{GPT-5.1}, all with medium reasoning effort. On $C_3$ instances (limited to 20 steps), all models demonstrated the capacity to break symmetry and reach the optimal 1-conflict state. Gemini 2.5 Pro achieved the optimal state by step 6, though it showed some instability after. Claude 4.5 Sonnet similarly converged to the optimal solution.

Most notably, GPT-5.1 not only broke symmetry but discovered a classic distributed systems technique: using unique node identifiers to resolve conflicts. The explicit strategy given by an agent goes as follows: 
\begin{displayquote}\textit{"[NEW] when repeated all-same-color conflicts persist, use node\_id ordering (e.g., smallest id switches, larger ids stay) to decide who adapts."} 
\end{displayquote}
This demonstrates that more capable models can rediscover known algorithms (e.g., leader election or priority-based conflict resolution) purely from in-context reasoning. While these tests lack the statistical rigor of our full benchmark, they suggest that the emergence of coordination strategies is a general property of scaling reasoning capabilities across different model families.

\subsection{Limitations}
While LoopBench provides a focused testbed for reasoning capabilities, our current study has several limitations. First, the number of tested models is limited to a few representative classes; a broader survey including more diverse open-weights and proprietary models would strengthen the generalization of these findings. Second, LLMs are known to be sensitive to prompt phrasing. Although we used standardized, generic prompts, subtle variations could potentially alter performance, and we have not yet performed a comprehensive sensitivity analysis. Third, our reliance on proprietary API-based models introduces reproducibility challenges, as backend model updates could affect future results. Finally, passing the full interaction history causes prompt length to scale as $O(T^2)$, which, combined with the cost of running large swarms of reasoning models, limits the scale of our experiments to relatively small graphs and short horizons.

\section{Future Work}
Future iterations of LoopBench will address current limitations and expand the scope of distributed reasoning agents. 
First, we plan to enhance the agent architecture by moving beyond free-text notes to \textbf{structured strategy representations} (e.g., for example structured code that the successive agents can make changes to rather than expressing the strategy through words alone). This would improve interpretability and allow for automated analysis of strategy evolution. To support longer horizons and larger networks, we also aim to implement \textbf{hierarchical memory systems} that compress interaction history, addressing the context window scalability issue inherent in full-history prompts. This should allow us to address the limitations of this work further.

Second, we intend to explore \textbf{heterogeneous swarms} where agents utilize different models or specialized roles (e.g., stabilizers vs. explorers). Preliminary results suggest that diverse reasoning capabilities may enhance robustness against symmetry traps, and we aim to formalize how such specialization emerges.

Third, we seek to establish a theoretical framework connecting problem complexity to agent reasoning depth. This involves quantifying the Reasoning Gap through information-theoretic metrics and exploring connections to active inference, moving towards a more formal understanding of how meta-cognitive reasoning enables distributed coordination. Finally, we will extend LoopBench to a broader class of distributed constraint satisfaction problems and real-world network scenarios, testing the generalizability of emergent LLM-based algorithms beyond graph coloring.

\section{Conclusion}
This work introduces \textbf{LoopBench}, a benchmark designed to evaluate the reasoning and coordination capabilities of LLM-based agents through the lens of distributed symmetry breaking. By focusing on the challenging problem of coloring over-constrained odd cycles, we create a minimal yet rigorous environment where simple heuristics fail, allowing the reasoning abilities of different models to be clearly distinguished.

Our experiments demonstrate that advanced reasoning models like O3 can successfully navigate these challenges, developing strategies to break deadlocks and achieve optimal solutions. This highlights the role of meta-cognitive reasoning in distributed problem solving, as agents must critique their own interaction history to synthesize policies that break symmetry. The analysis of their strategies offers a unique, interpretable view into the emergence of these collective behaviors.

The findings highlight a measurable "Reasoning Gap" between different classes of models and underscore the potential of LLMs as components in sophisticated multi-agent systems. LoopBench provides a standardized framework for studying this gap and for exploring the principles of emergent intelligence in language-based agent swarms. Future work will expand the benchmark with a wider range of models and explore extensions to more complex distributed problems, advancing toward more capable and coordinated AI systems.

\bibliographystyle{splncs04}
\bibliography{references_ANTS}
\begin{comment}
IEEE conference templates contain guidance text for composing and formatting conference papers. Please ensure that all template text is removed from your conference paper prior to submission to the conference. Failure to remove the template text from your paper may result in your paper not being published.

\end{comment}
\end{document}